\documentclass[runningheads]{llncs}
\usepackage{graphicx}
\pdfoutput=1
\usepackage{tikz}
\usepackage{comment}
\usepackage{graphicx}
\usepackage{amsmath,amssymb} 
\usepackage{color}
\usepackage{multirow}
\usepackage{booktabs}
\usepackage{float}
\usepackage{subfigure}
\usepackage{wrapfig}
\usepackage[marginal]{footmisc}

\usepackage[accsupp]{axessibility}  

\begin{document}
\pagestyle{headings}
\mainmatter
\def\ECCVSubNumber{3600}  

\title{SIT: A Bionic and Non-Linear Neuron for Spiking Neural Network} 



\author{Cheng Jin \inst{1}\orcidID{0000-0002-3522-3592
} \and
Rui-Jie Zhu\inst{2}\orcidID{0000-0003-4864-8474}
\and Xiao Wu\inst{3}\orcidID{0000-0002-1259-8674}\and
Liang-Jian Deng\inst{3}\orcidID{0000-0003-3178-9772
}}
\authorrunning{C. Jin et al.}
%
\institute{School of Optoelectronic Science and Engineering, University of Electronic Science and Technology of China\\ \email{Cheng.Jin@std.uestc.edu.cn}\\\and
School of Public Affairs and Administration, University of Electronic Science and Technology of China\\
\email{ridger@std.uestc.edu.cn}\\ \and
School of Mathematical Sciences, University of Electronic Science and Technology of China\\
\email{wxwsx1997@gmail.com, liangjian.deng@uestc.edu.cn}}
\maketitle
\footnote{The first author and second author contribute equally to this work.\\}
\begin{abstract}
Spiking Neural Networks (SNNs) have piqued researchers' interest because of their capacity to process temporal information and low power consumption. However, current state-of-the-art methods limited their biological plausibility and performance because their neurons are generally built on the simple Leaky-Integrate-and-Fire (LIF) model. Due to the high level of dynamic complexity, modern neuron models have seldom been implemented in SNN practice. In this study, we adopt the Phase Plane Analysis (PPA) technique, a technique often utilized in neurodynamics field, to integrate a recent neuron model, namely, the Izhikevich neuron. Based on the findings in the advancement of neuroscience, the Izhikevich neuron model can be biologically plausible while maintaining comparable computational cost with LIF neurons. By utilizing the adopted PPA, we have accomplished putting neurons built with the modified Izhikevich model into SNN practice, dubbed as the Standardized Izhikevich Tonic (SIT) neuron. For performance, we evaluate the suggested technique for image classification tasks in self-built LIF-and-SIT-consisted SNNs, named Hybrid Neural Network (HNN) on static MNIST, Fashion-MNIST, CIFAR-10 datasets and neuromorphic N-MNIST, CIFAR10-DVS, and DVS128 Gesture datasets. The experimental results indicate that the suggested method achieves comparable accuracy while exhibiting more biologically realistic behaviors on nearly all test datasets, demonstrating the efficiency of this novel strategy in bridging the gap between neurodynamics and SNN practice.

\end{abstract}

\section{Introduction}

\begin{figure}
    \centering
    \includegraphics[width=8cm]{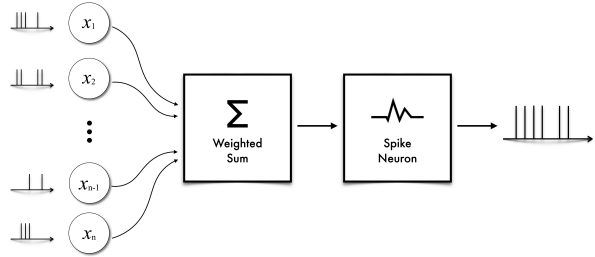}
    \caption{General schematic diagram of Spiking Neural Network (SNN). Different from ANNs, in SNNs, the inputs are encoded and propagates as a series of spikes.}
    \label{fig:flow}
\end{figure}

\begin{wrapfigure}{R}{0.5\textwidth}
\begin{center}
    \includegraphics[width=0.48\textwidth]{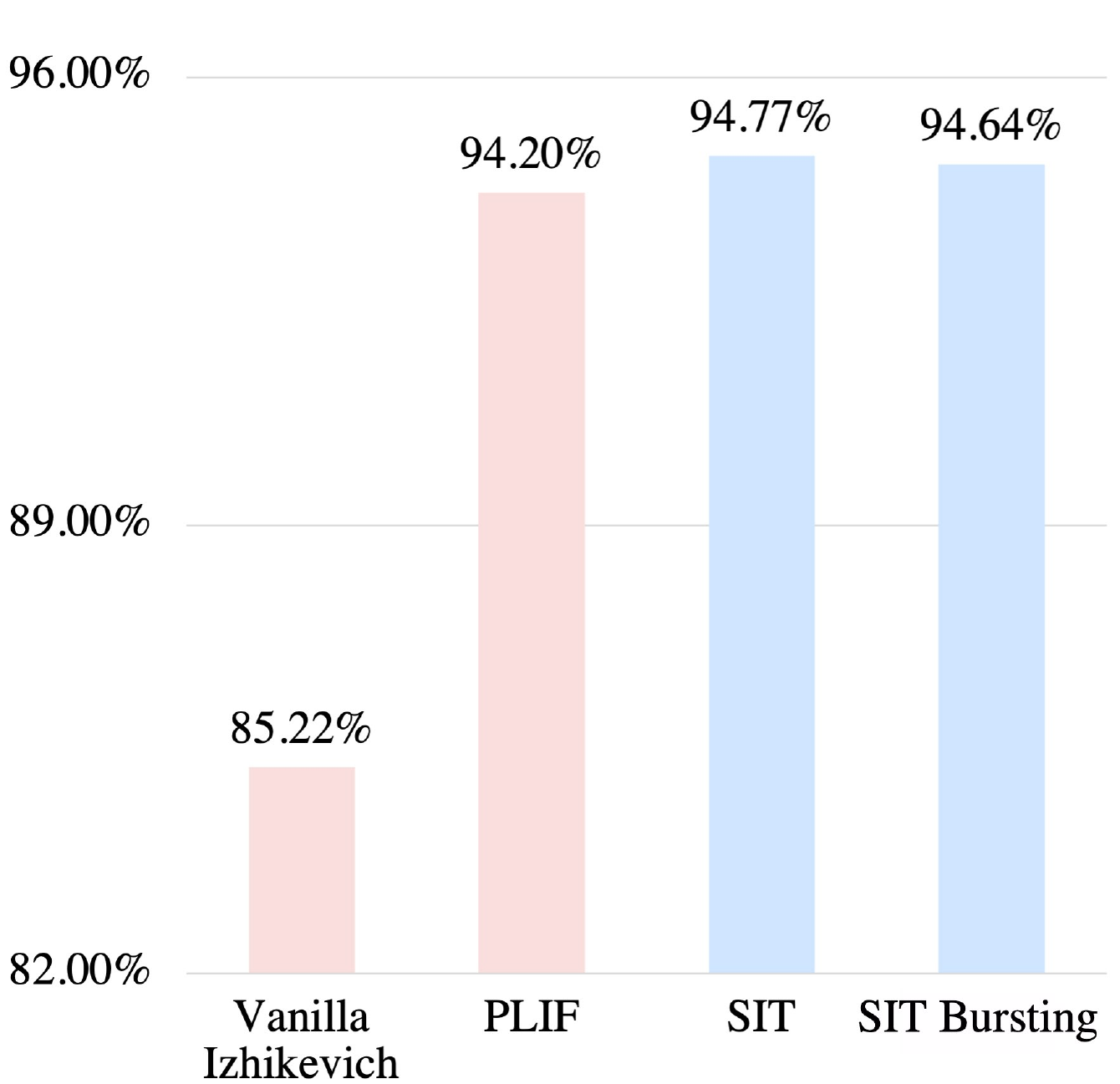}
\end{center}
    \caption{Classification accuracy in Fashion-MNIST dataset with convolutional spike neural network based on different neurons.}
    \label{fig:contrast}
\end{wrapfigure}
Being considered as the thrid generation of neural network models \cite{maass1997networks}, Spiking Neural Networks (SNNs) have a greater biological interpretability. This draws a lot of attention from academics because of its unique qualities, such as excellent biological plausibility \cite{gerstner2014neuronal}, low power consumption \cite{roy2019towards}, and compatibility with event-driven datasets. As illustrated in Fig. \ref{fig:flow}, the information flown in SNNs is as a form of spike sequence. By completing this art of transformation, SNN can approximately complete the interference similar to biology mechanism. Additionally, the application potential of SNN algorithms has prompted several commercial firms to develop neuromorphic computer chips capable of running SNN algorithms at a substantially lower energy cost \cite{akopyan2015truenorth}. Recently, the Leaky-Integrate-and-Fire (LIF) model~\cite{lif04} is proposed to address spike-frequency adaptation in temporal dynamics. In ReSuMe~\cite{ponulak2010supervised},
LIF acts a explicitly iterative neuron as the part of model to train. To get different types of neurons, PLIF~\cite{fang2021incorporating} uses learnable parameters to associate neuron membrane to learn faster and better than previous SNNs. 

However, with the development of computational neuroscience, generally, neuron models can be divided into two parts from the perspective of whether they have an adaptive reset mechanism or not. Integrate-and-Fire (IF), Leaky-Integrate-and-Fire (LIF) \cite{koch1998methods}, and Quadratic-Integrate-and-fire (QIF) \cite{gerstner2014neuronal} models represent the non-adaptive-reset categories. On the contrary, the Hodgkin–Huxley \cite{hodgkin1952quantitative} and Izhikevich \cite{izhikevich2003simple} have adaptive resetting settings, which is more biologically plausible.  The Izhikeivch neuron has been utilized in Field Programmable Gate Array (FPGA) \cite{rice2009fpga} and use the Spike-Timing-Dependent Plasticity (STDP) \cite{stdpizhikevich01,stdpizhikevich02} for training. Regrettably, the attempt of integrating Izhikeivch neuron into spike-based back-propagation fails because of the high complexity in neurodynamics \cite{machado2019natcsnn}, which would lead to unsatisfactory performance, as Fig. \ref{fig:contrast} shows. To be more precisely, the update rule of the Izhikevich neuron contains several dimensionless parameters in order to simulate the different neuron behaviors, which requires extra knowledge to integrate to neural networks. Phase Plane Analysis (PPA) is widely used in neurodynamics. Accroding to \cite{izhikevich2007dynamical}, several neurons, such as LIF, IF, QIF and the Izhikevich neurons are analyzed through PPA to dig into their relationship between biological behavior and mathematics.

In this paper, we accomplish the integration by rethinking from the neurodynamics field and combine the idea of normalization in neural networks to qualitatively determine the dimensionless parameters and propose a network architecture to verify our hypothesis.

The main contributions of this paper can be summarized as follows: 
\begin{enumerate}
	\item[1)]
	We adopt Phase Plane Analysis (PPA) technique to integrate the Izhikevich model into SNN practice. Furthermore, the PPA technique has the potential to introduce other recent non-linear neuron into SNNs, such as Exponential Integrate-and-Fire (EIF) neuron \cite{eif}, Adaptive Exponential Integrate-and-Fire (AdEx) neuron \cite{adex} and LIF with Spike-frequency adaptation \cite{2001lifadaption}.
	\item[2)] We employ PPA to adapt the Izhikeivch neuron in Tonic Spiking and Bursting settings into Standardized Izhikevich Tonic (SIT) neuron. To reduce the computational consumption and introduce variability of SNN, we mix the LIF neuron and SIT neuron in separate convolutional layers, and ultimately propose the Hybrid Neuron Network (HNN).
	\item[3)] We validate that the neuron with higher dynamic complexity could achieve state-of-the-art performance in some area by testing our approaches on both traditional static MNIST \cite{lecun1998mnist} and Fashion-MNIST \cite{xiao2017fashionmnist} benchmarks, which are frequently used in validating the performance of ANNs, and N-MNIST \cite{orchard2015nmnist}, CIFAR10-DVS  \cite{li2017cifar10dvs}, DVS128 Gesture \cite{amir2017dvsg} neuromorphic datasets.
\end{enumerate}	

\section{Related Works and Motivation}
\subsection{Leaky-Integrate-and-Fire (LIF) Neuron Model}
Proposed by Koch \textit{et al.} \cite{koch1998methods}, the Leaky-Integrate-and-Fire (LIF) neuron model performs as an integrator. It accumulates the membrane voltage by receiving the stimulus of external input current in a simple differential equation:
\begin{equation}\label{lif equation}
    \tau\frac{dU(t)}{dt} = -(U(t)-U_{r}) + I(t),
\end{equation}
where $U(t)$ represents the membrane potential of the neuron at time $t$, $I(t)$ represents the input to neuron at time $t$, and $\tau$ is the membrane time constant. The equation describing the update of voltage before spiking is dubbed as subthreshold dynamics. When $U(t)$ reaches the threshold, the neuron would produce a spike, and $U(t)$ would be reset to $U_{r}$. Currently, LIF has been implemented into SNN and can be trained 
using STBP~\cite{lif01,lif03} and Surrogate Gradient and achieves state-of-the-art accuracy~\cite{fang2021incorporating}.

\subsection{Izhikevich Neuron Model}\label{sec:izhikevich model}
The Izhikevich model is a quadratic integrate-and-fire neuron with reference recovery variables. Several spiking neuron models have been developed by neuroscientists to accurately describe the interactions between the biological neuron's input and output signals. Among the proposed neuron models, the Izhikevich model is both biologically reasonable and computationally inexpensive. However, as the introduction of self-adaptation mechanism, The Izhikevich neuron's sub-threshold dynamics are formed by a set of differential equation defined as:

\begin{equation}
\begin{aligned}
    \tau\frac{\mathrm{d}U(t)}{\mathrm{d}t} &= 0.04U(t)^2 +5V + 140 -U(t) + I(t),\\
	\tau\frac{\mathrm{d}V(t)}{\mathrm{d}t} &= a(bU(t)-V(t)),
	\label{Izhikevich dynamics}
\end{aligned}
\end{equation}
with the auxiliary after-spike resetting:
\begin{equation}
\begin{aligned}
    \text{if} \quad &U(t) \ge 30 \text{mV}, \text{then}\\ 
    &U(t+\frac{1}{\tau}) = c,\\ 
    &V(t+\frac{1}{\tau}) = V(t)+d,
    \label{Izhikevich resetting}
\end{aligned}
\end{equation}
where $U(t)$ represents the membrane potential of the neuron at time $t$, $X(t)$ represents the input to neuron at time step $t$,  and $\tau$ is the membrane time constant. The \textit{a,b,c,d} are the hyper-parameters of the Izhikevich neuron: \textit{a} represents the time scale of recovery variable $V$,
\textit{b} is the sensitivity of $V$ to membrane potential, \textit{c} shows the voltage to reset to after a spike, \textit{d} plays for resetting the recovery value after a spike. In the scenario of the Izhikevich settings, 30 mV is the threshold of spiking and the spiking reset voltage is -65 mV. In the original model \cite{izhikevich2003simple}, the author has proposed up to 20 neuron models to mimic the actual biological behavior. Among them, \emph{tonic} spiking holds the default setting of the model, since most spiking type of biological neurons falls on tonic spiking, in which $a=0.02, b=0.2, c=-65, d=8$. However, Izhikevich model has different neuron behaviors and intricated updating rules, which makes it very difficult to bridge the gap between Izhikevich model and SNN.

\subsection{Spike-based Back Propagtion}
The discussions of Spike-based back propagation techniques should begin with SpikeProp \cite{bohte2002error}. To get around the non-differentiable threshold-triggered firing mechanism, SpikeProp employs spike response model (SRM) and uses the linear approximation. However, the SpikeProp could only be applied to single-layer SNNs. To introduce the back-propagation into multi-layer SNNs and achieve better performance, the approach called Surrogate Gradient was proposed, which could directly be applied by spiking neurons and back-propagate by unfolding the network through time steps \cite{surrogatelif}. 

\subsection{Phase Plane Analysis}
Phase Plane Plot is a geometric representation of the trajectories of a two-dimensional dynamical system in the phase plane, which is widely used in neurodynamics analysis. Its analysis process contains the calculation of the eigenvalue and the eigenvector of the derivative equation systems, following by plotting the eigenvectors in the vector field of the derivative equations. By observing the phase plane plot, one can identify the possible stable points and the unstable points of the system.

In order to depict the Phase Plane Plot, we summarize the following steps:
\begin{itemize}
    \item Plot the vector field of the differential equation system.
    \item Plot the nullcline of each variable in the differential equation, and calculate the coordinates of the intersection point.
    \item Calculate the eigenvalues and eigenvectors of the differential equations.
    \item Based on the positive and negative of the eigenvalues, determine the stability of the intersection point.
\end{itemize}

\subsection{Motivation}
Neurons are crucial in determining the performance of SNNs. It has been established in neurodynamics that the Izhikevich neuron has more biological plausibility while maintaining low computational costs. This highlights the importance of incorporating various neuron types into Spiking Neural Networks. However, the vanilla Izhikevich model's performance is not guaranteed (full experiment results are shown in Fig. \ref{fig:contrast}). As a result, we contemplate analyzing the success of the Leaky-Integrate-and-Fire (LIF) neurons and modifying the Izhikevich model to get satisfactory outcomes. Recalling the differential description of LIF Neurons, as Eq. \ref{lif equation} indicates:

\begin{figure}[H]
    \centering
    \includegraphics[width=\linewidth]{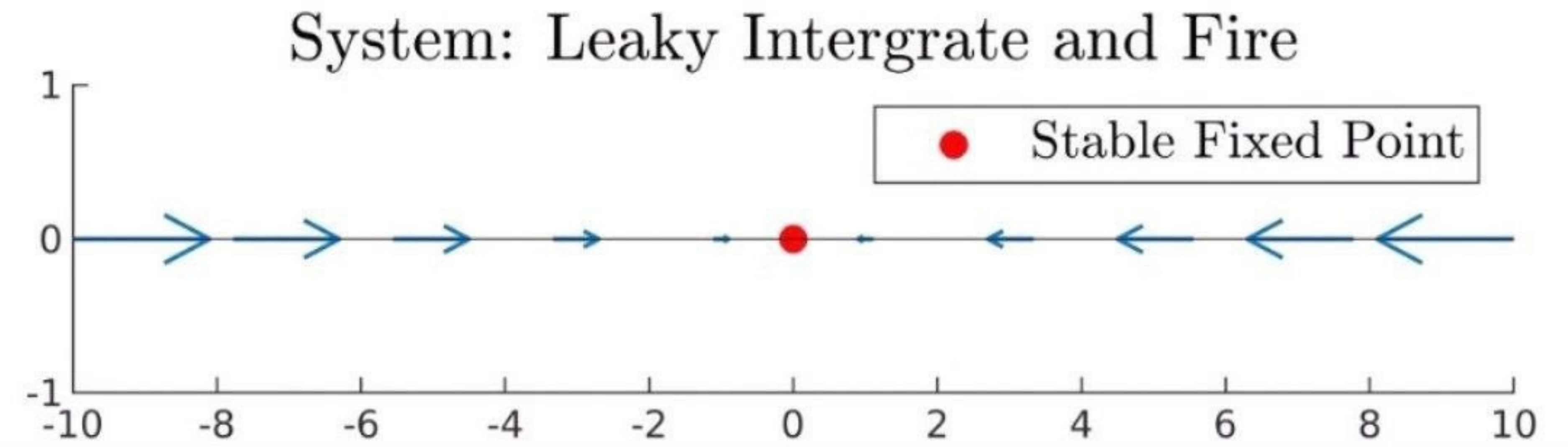}
    \caption{Vector field of the sub-threshold expression of the Leaky-Integrate-and-Fire (LIF) neurons.}
    \label{fig:LIF_plot}
\end{figure}

\begin{figure}[H]
    \begin{minipage}[t]{0.5\linewidth}
    \centering
    \includegraphics[width=1\linewidth]{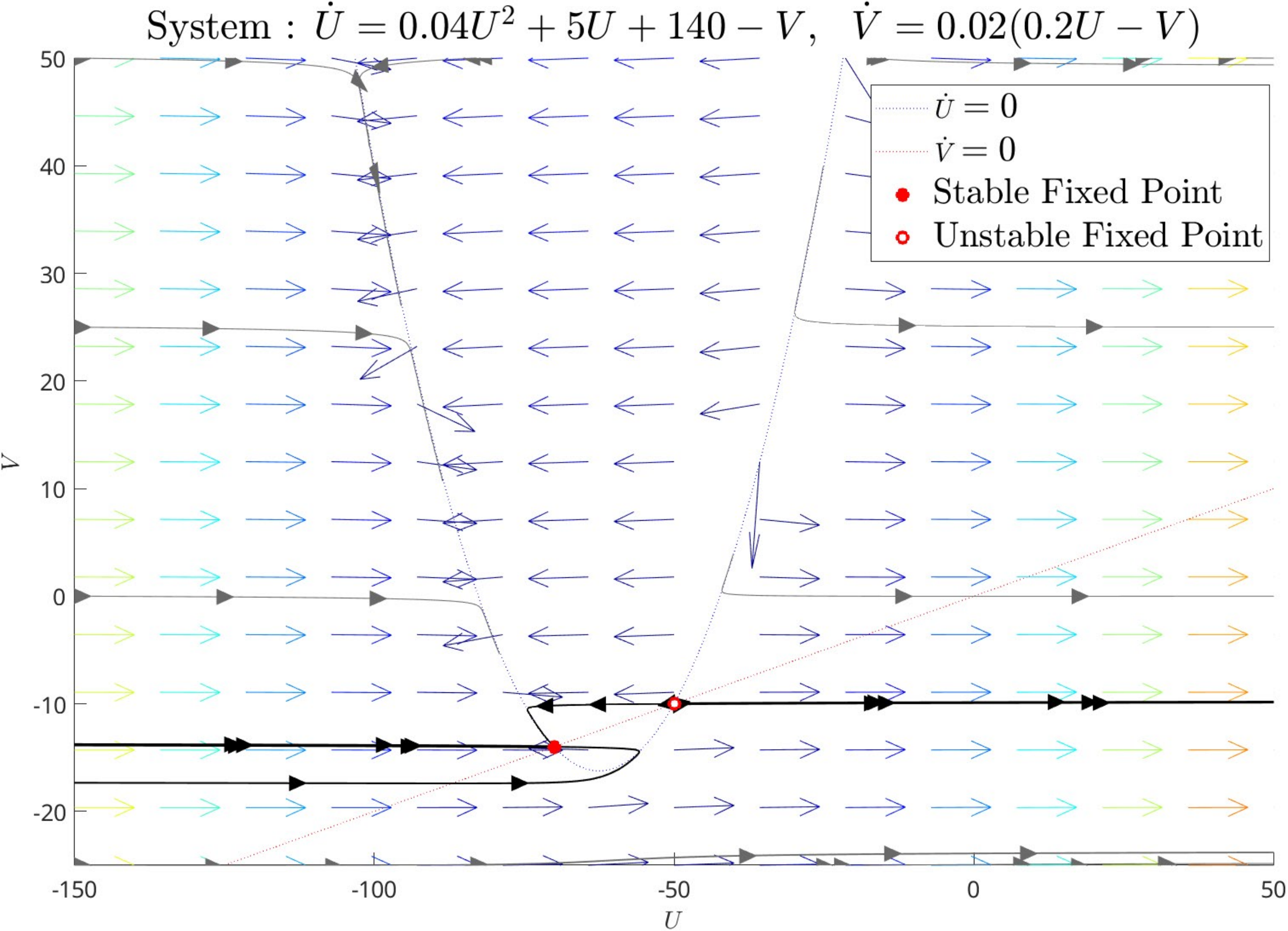}
    \caption{Phase plane plot of the vanilla Izhikevich model in Tonic spiking settings.}
    \label{fig:orig_izhikevich}
    \end{minipage}
    \begin{minipage}[t]{0.5\linewidth}
    \centering
    \includegraphics[width=1\linewidth]{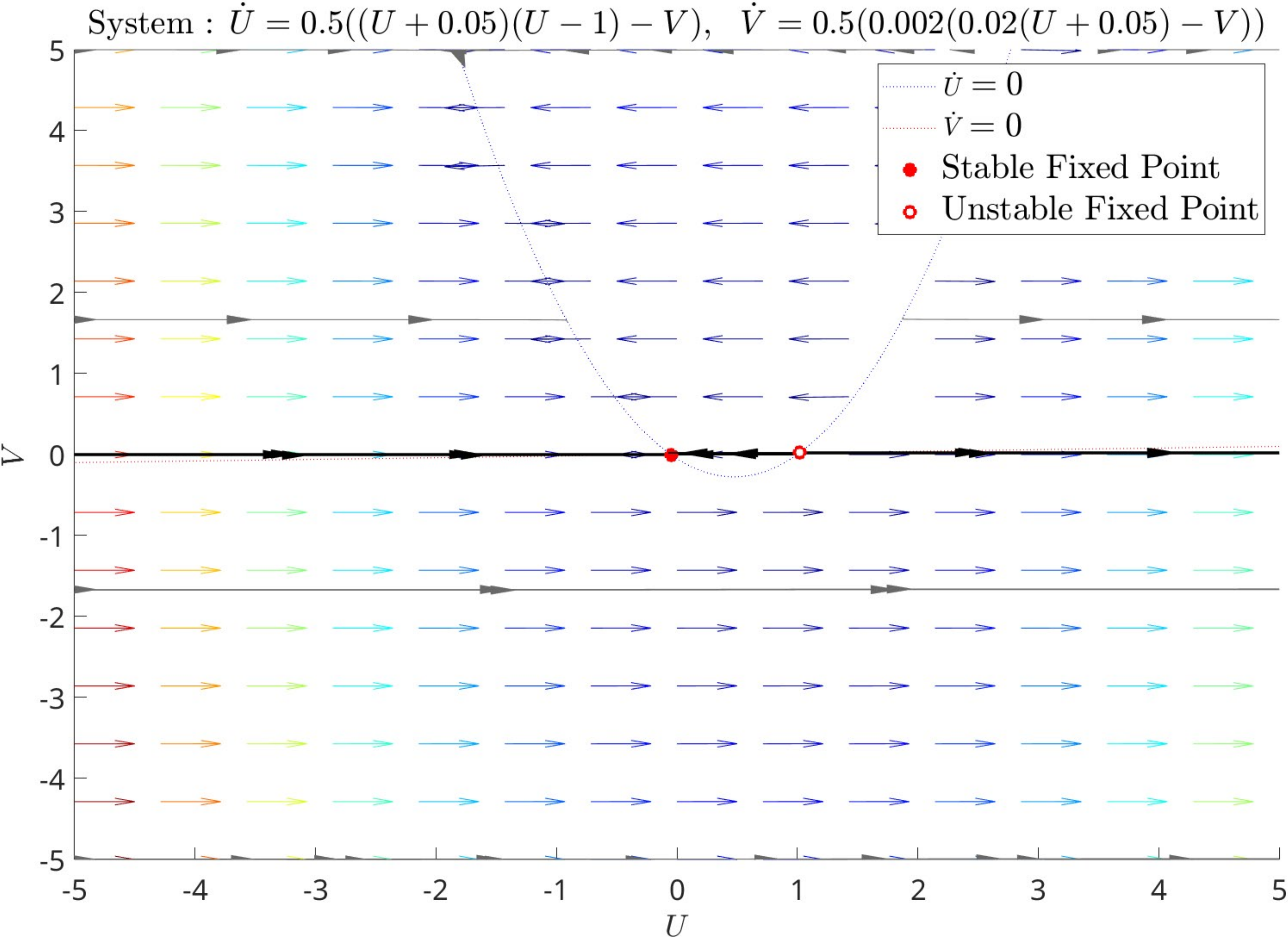}
    \caption{Standardized Izhikevich model. The stable point is at $(-0.05, 0)$, and the unstable point is at $(1.02, 0.021)$, ensuring the forward propagation process and spiking.}
    \label{fig:std}
    \end{minipage}
\end{figure}

Due to the system being in one-dimensional form, we can only plot the vector field of this system. It is clear that regardless of the initial state, the system always converges in a stable direction, \textit{i.e.}, each input serves as the initial condition for this differential equation. By switching tonic settings and plotting the associated sub-threshold phase plane plot, as seen in Fig. \ref{fig:orig_izhikevich}, we can observe that the fixed stable point remains at $(-70, -14)$. However, because the initial pixel value inputs are often positive, it is difficult to spike and propagate useful features in the SNN. As a result, we are motivated to use PPA to demonstrate the internal neurodynamics of more sophisticated neuron models and to incorporate them into the current SNN framework in a targeted manner, \textit{i.e.}, to alter the Izhikevich neuron into its stable state.


\section{Method}
In this section, we first review the background in Sect. \ref{sec:model intro}, and introduce the Phase Plane Analysis (PPA) procedure in Sect. \ref{sec:PPA}. The Izhikevich model is then introduced in Sect. \ref{sec:model}. At last, we describe the learning algorithm of SNNs in Sect. \ref{sec:BP}.

\subsection{Preliminaries}\label{sec:model intro}
\subsubsection{The standardization of biological neuron models}
There are two major types of biological neuron models, the Leaky-Integrate-and-Fire (LIF) model and the Quadratic-Integrate-and-fire (QIF) model, the sub-threshold neuronal dynamics equation of LIF is illustrated in Eq. \ref{lif equation}. As a quadratic neuron model, the QIF model replaces the linear term with the quadratic term. The sub-threshold expression of QIF model is shown in Eq. \ref{qif equation}.
\begin{equation}\label{qif equation}
    \tau\frac{dU(t)}{dt} = k(U(t) - U_{r})(U(t) - U_c) + I(t).
\end{equation}
In Eq. \ref{qif equation}, the hyper-parameter $U_c$ can be understood as the critical voltage for a brief current pulse to initiate a spike, and $U_{r}$ represents the simulating reset voltage, which is close to the real reset voltage $U_{reset}$.
Comparing with Eq. \ref{qif equation}, we may rewrite the Izhikevich model as Eq. \ref{std iz} denotes. When the $U(t)$ hits the $U_{threshold}$, the reset process is described in Eq. \ref{Izhikevich resetting}. In this form, we may consider that the Izhikevich model is a special form of the QIF model, it introduces a recovery variable $V$. 
\begin{equation}\label{std iz}
    \begin{aligned}
        \tau \frac{\mathrm{d}U(t)}{\mathrm{d}t} &= k(U(t)-U_{r})(U(t)-U_{c}) -V(t) + I(t),\\
        \tau \frac{\mathrm{d}V(t)}{\mathrm{d}t} &= a(b(U(t)-U_{r})-V(t)).
    \end{aligned}
    \end{equation}

 For better biological plausibility, the $U_{reset}$  is set to around -65 $mV$ and the $U_{threshold}$ is set to around 10 $mV$ in early SNN simulation frameworks \cite{hazan2018bindsnet}. However, 
 the derivative of the threshold-triggered firing mechanism is defined by surrogate derivatives \cite{fang2021incorporating,fang2020iir,kaiser2020decolle,wu2019neunorm,zhang2019strsbp}, and Surrogate Gradient is defined with surrogate derivatives and its value range is [0, 1]. Based on that, the neuron of spike-based back propagation need to normalize its membrane potential to [0, 1] during training process, which means the $U_{reset}$ of neuron should be set to 0 and the $U_{threshold}$ need be set to 1. However, the built-in dimensionless hyper-parameter $a,b,c,d$ should be recalculated and determined for the change of threshold and reset voltage of Izhikevich model. 

\subsection{The PPA Procedure}\label{sec:PPA}

\begin{theorem}
The nullclines of the Standardized Izhikevich model in Phase Plane Plot have two intersections. One of the intersections is the stable point of the Izhikevich model.
\end{theorem}
\textit{Proof.}

By the standardized form of the Izhikevich model, we firstly investigate the property of the intersection of the nullclines. By substituting $C_m \frac{dU}{dt}=0, \frac{dV}{dt}=0$, we have:

\begin{equation}\label{eq:derivation_1}
	\begin{aligned}
		0&=k\left(U-U_{r}\right)\left(U-U_{c}\right)-V+I, \\
		0&=a\left(b\left(U-U_{r}\right)-V\right),\\
		&\Longrightarrow V=b(U-U_{r}).
	\end{aligned}
\end{equation}

Bringing $V=b(U-U_{r})$ back to the initial equations, we would have:
\begin{equation}\label{eq:quadratic}
	\begin{aligned}
        & kU^2-[b+k(U_{r}+U_{c})]U+(I-bU_r-kU_rU_{c})=0,\\
        \Longrightarrow &U=\frac{b+k\left(U_{r}+U_{c}\right) \pm \sqrt{\left(b+k\left(U_{r}+U_{c}\right)\right)^{2}-4 k\left(I-b U_{r}-k U_{r} U_{c}\right)}}{2 k},
	\end{aligned}
\end{equation}

and if $\Delta>0$, the negative solution falls on:
\begin{equation}\label{eq:negative_solution}
	U_-=\frac{b+k\left(U_{r}+U_{c}\right) - \sqrt{[b+k(U_{c}-U_{r})]^2-4Ik}}{2 k}.
\end{equation}

With that, the range of parameter $b$ can be determined after the voltage parameters $U_r, U_c$ and the experience parameter $k$ is settled.

With this expression, we can determine the coordinates of the stable point by changing parameter $b, k, U_r, U_{c}$. 

When $\Delta=0$, $U_+$ and $U_-$ coincides together, at this moment:

\begin{equation}
	\begin{aligned}
	\sqrt{[b+k(-U_{r}+U_c)]^{2}-4 k I}=0, \\
	\Longrightarrow I=\frac{[b+k(-U_{r}+U_{c})]^{2}}{4 k}.
	\end{aligned}
\end{equation}

Bringing this $I$ into original solution, Eq. \ref{eq:quadratic}, we would have the coincide voltage $U_o$:

\begin{equation} \label{eq:voltage}
U_o=\frac{b+k(U_{r}+U_{c})}{2k}>U_-.
\end{equation}

The rest of delta conditions would cause occurrence of unstable and saddle point, and this part can cause resonator property of the Izhikevich model, which is out of cover in this paper. Interested readers may refer to \cite{izhikevich2007dynamical} for detail.

Next, we would proof the stability of this point. According to dynamical system analysis, a point is stable if and only if its corresponding eigenvalues are all negative. Here we will proof the negativity of the eigenvalues.

We firstly derive the eigenvalues of the Jacobian matrix based on the original differential equations:

\begin{equation}
	J=\left(\begin{array}{ll}
		\frac{\partial \frac{d U}{d t}}{\partial U} & \frac{\partial \frac{d U}{d t}}{\partial V} \\
		\frac{\partial \frac{d V}{d t}}{\partial U} & \frac{\partial \frac{d V}{d t}}{\partial V}
	\end{array}\right),
\end{equation}

in which:

\begin{equation}
	\begin{aligned}
		\frac{\partial \frac{d U}{d t}}{\partial U}&=\frac{\partial}{\partial V} \frac{k\left(U-U_{r}\right)\left(U-U_{c}\right)-V+I}{C_m}=\frac{2 U-k\left(U_{r}+U_{c}\right)}{C_{m}}, \\
		\frac{\partial \frac{d U}{d t}}{\partial V}&=\frac{\partial}{\partial V} \frac{k\left(U-U_{r}\right)\left(U-U_{c}\right)-V+I}{C_m}=-\frac{1}{C_{m}}, \\
		\frac{\partial \frac{d V}{d t}}{\partial U}&=\frac{\partial}{\partial U} a\left(b\left(U-U_{r}\right)-V\right)=a b, \\
		\frac{\partial \frac{d V}{d t}}{\partial V}&=\frac{\partial}{\partial V}, a\left(b\left(U-U_{r}\right)-V\right)=-a.\\
		\Longrightarrow &J=\left(\begin{array}{ll}
		\frac{2 U-k\left(U_{r}+U_{c}\right)}{C_{m}} &  -\frac{1}{C_{m}} \\
		ab & -a
	\end{array}\right)
	\end{aligned}
\end{equation}

The corresponding eigenvalues of this matrix can be solved as:

\begin{equation}\label{eq:lambda}
	\begin{aligned}
	&\lambda=-a-\frac{k}{C_{m}}\left(U_{r}+U_{c}-2 U\right) \\
	&\pm \sqrt{\left(a-\frac{k}{C_{m}}\left(U_{r}+U_{c}-2 U\right)\right)^{2}-4\left[a \frac{b}{C_{m}}+a \frac{k}{C_{m}}\left(U_{r}+U_{c}-2 U\right)\right]}.
	\end{aligned}
\end{equation}

Because we are investigating the $U_-$ point, substitute $U$ with $U_{-}$ in Eq. \ref{eq:lambda}. Initially, we would take care of the outside the radical, \textit{i.e.}, suppose $-a-\frac{k}{C_{m}}\left(U_{r}+U_{t h}-2 U\right)<0$, due to biological constrain, $a>0$, So $-a$ is always negative. Therefore what we want to proof is that:

\begin{equation}
    \begin{aligned}
    U_{r}+U_{c}-2 U_-<0 \\
    \Longrightarrow U_{r}+U_{c}<2 U_-.
    \end{aligned}
\end{equation}

Recall Eq. \ref{eq:voltage}, $U_o=\frac{b+k(U_r+U_{th})}{2k}>U_-$. With biological constrain $b, k>0$, collate the expression, we would see:

\begin{equation}
    \begin{aligned}
    &\frac{b+k(U_{r}+U_{c})}{2k}>U_-\\
    \Longrightarrow &\frac{b}{k}+(U_{r}+U_{c})>2U_->U_{r}+U_{c}.
    \end{aligned}
\end{equation}

Therefore, $-a-\frac{k}{C_{m}}\left(U_{r}+U_{c}-2 U\right)<0$ is proofed, and this ensured at least one eigenvalue is negative. 

Moving into the radical, suppose $b=0$, Eq. \ref{eq:lambda} would fall into:

\begin{equation}
    \begin{aligned}
	&\lambda=-a-\frac{k}{C_{m}}\left(U_{r}+U_{c}-2 U_{-}\right)+\sqrt{\left(a+\frac{k}{C_{m}}\left(U_{r}+U_{c}-2 U_{-}\right)\right)^{2}}\\
	&=0.
\end{aligned}
\end{equation}

which is contradicted with the delta condition. Actually, under biological constraint, $b>0$, thus ensuring the negativity of all radical value. Therefore, the two eigenvalues are negative, ensuring the stability of $U_-$ point.

\begin{figure}[H]
    \centering
    \includegraphics[width=0.8\linewidth]{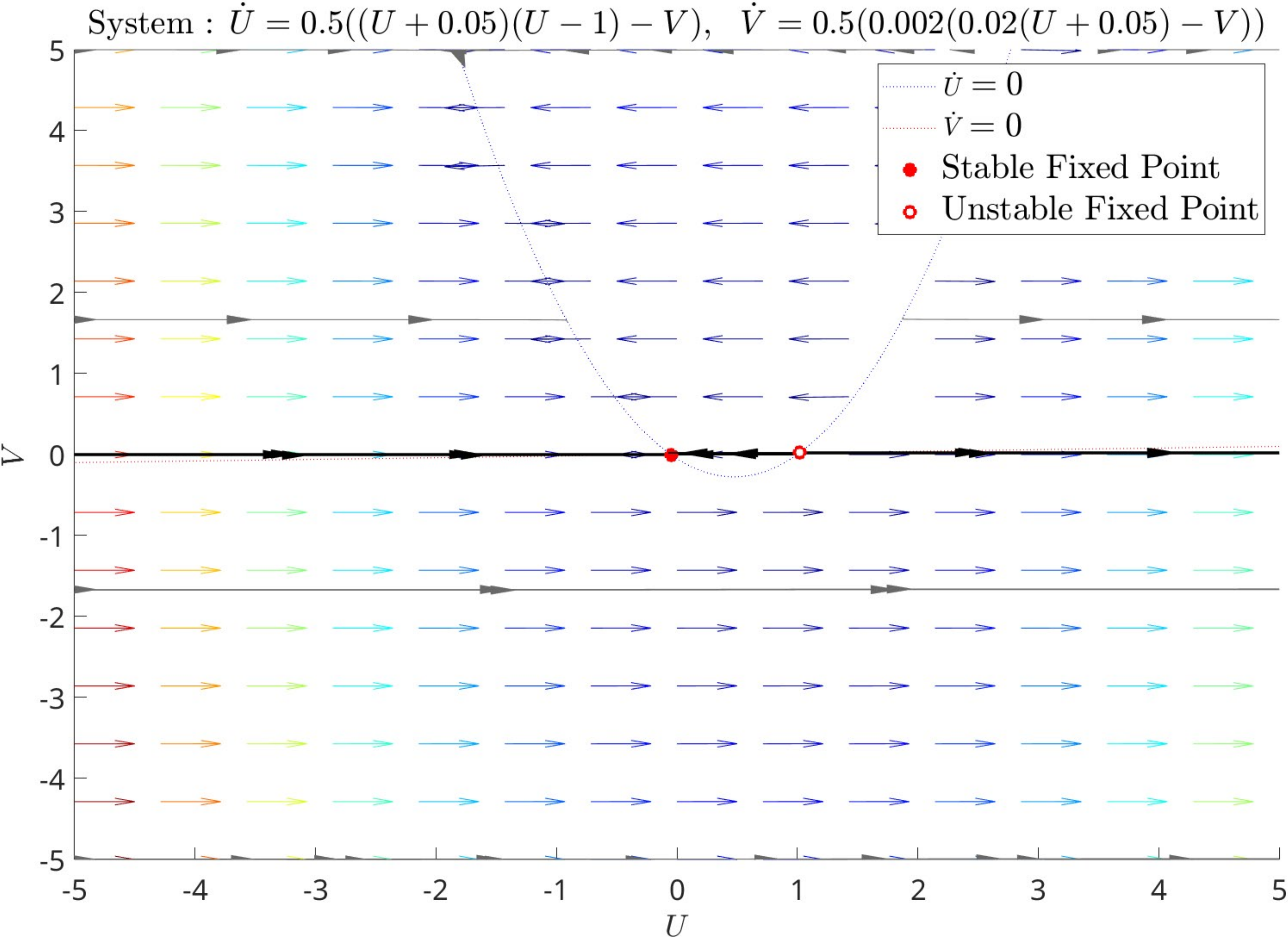}
    \caption{Standardized Izhikevich model. The stable point is at $(-0.05, 0)$, and the unstable point is at $(1.02, 0.021)$, which ensures the forward propagation process and spiking.}
    \label{fig:std}
\end{figure}

\subsection{Standardization of the Izhikevich Model}\label{sec:model}
The failure of the original model can be explained after the PPA procedure. The original tonic spiking parameter's PPA outcome suggests that the original model's stable point falls on $(-70, -14)$. However, regular inputs of the neural network would be pre-processed to standardized them within a positive range, particularly in [0,1]. Recall Eq. \ref{Izhikevich dynamics}, such inputs would have a negligible effect on the firing of the neurons, resulting in the propagation of undifferentiated signals. As a result, we are inspired by the concept of normalization, which compresses the inputs into a smaller range and accelerates network convergence \cite{ba2016layer,wu2018group}, and offer our strategy aimed at getting the stable point within the inputs' normal range.

\begin{figure}[H]
    \begin{minipage}[t]{0.5\linewidth}
    \centering
    \includegraphics[width=1\linewidth]{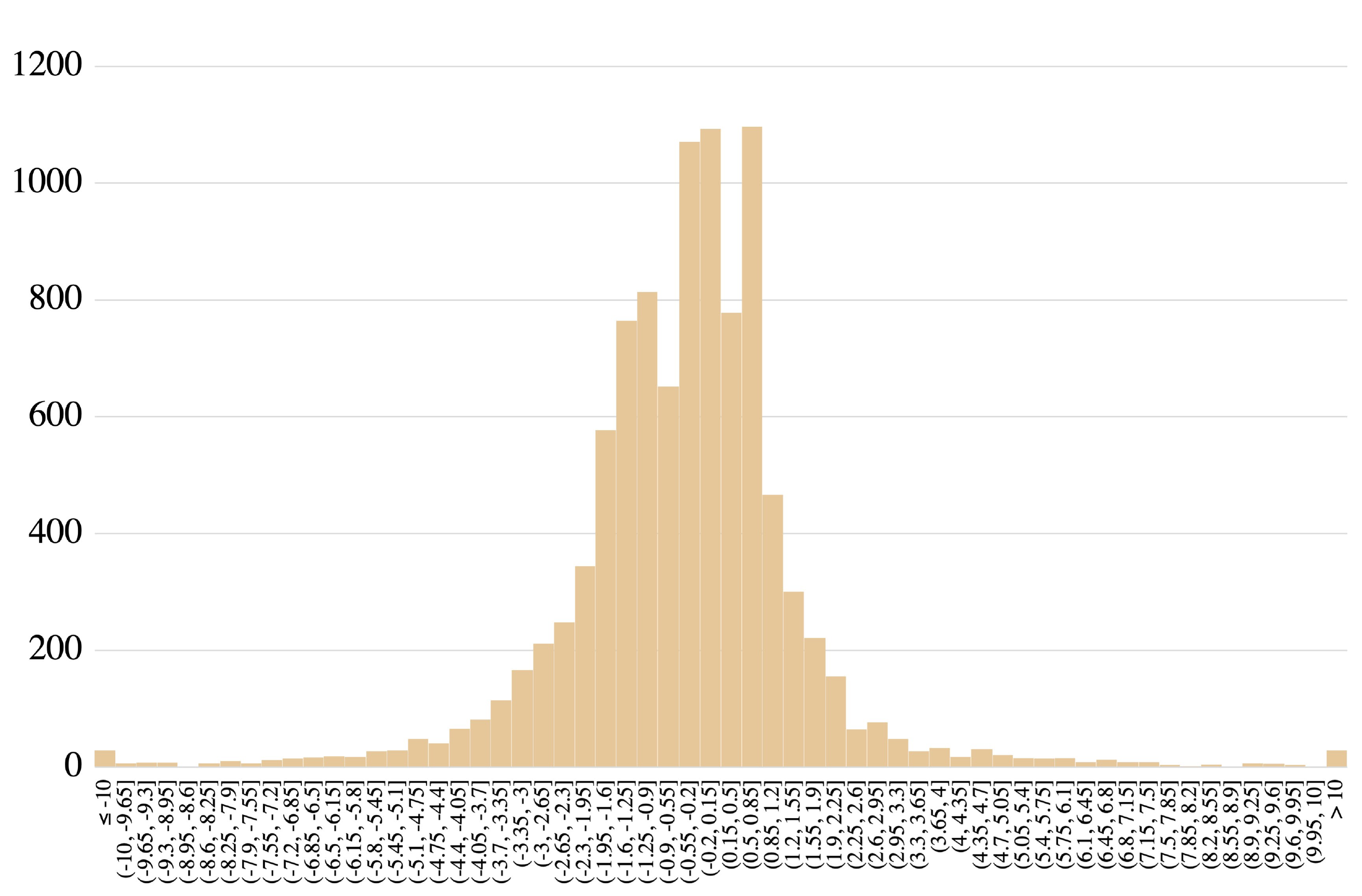}
    \caption{Output distribution of spike encoding convolutional layer, sampling 10,000 samples. }
    \label{fig:distribution}
    \end{minipage}
    \begin{minipage}[t]{0.5\linewidth}
    \centering
    \includegraphics[width=1\linewidth]{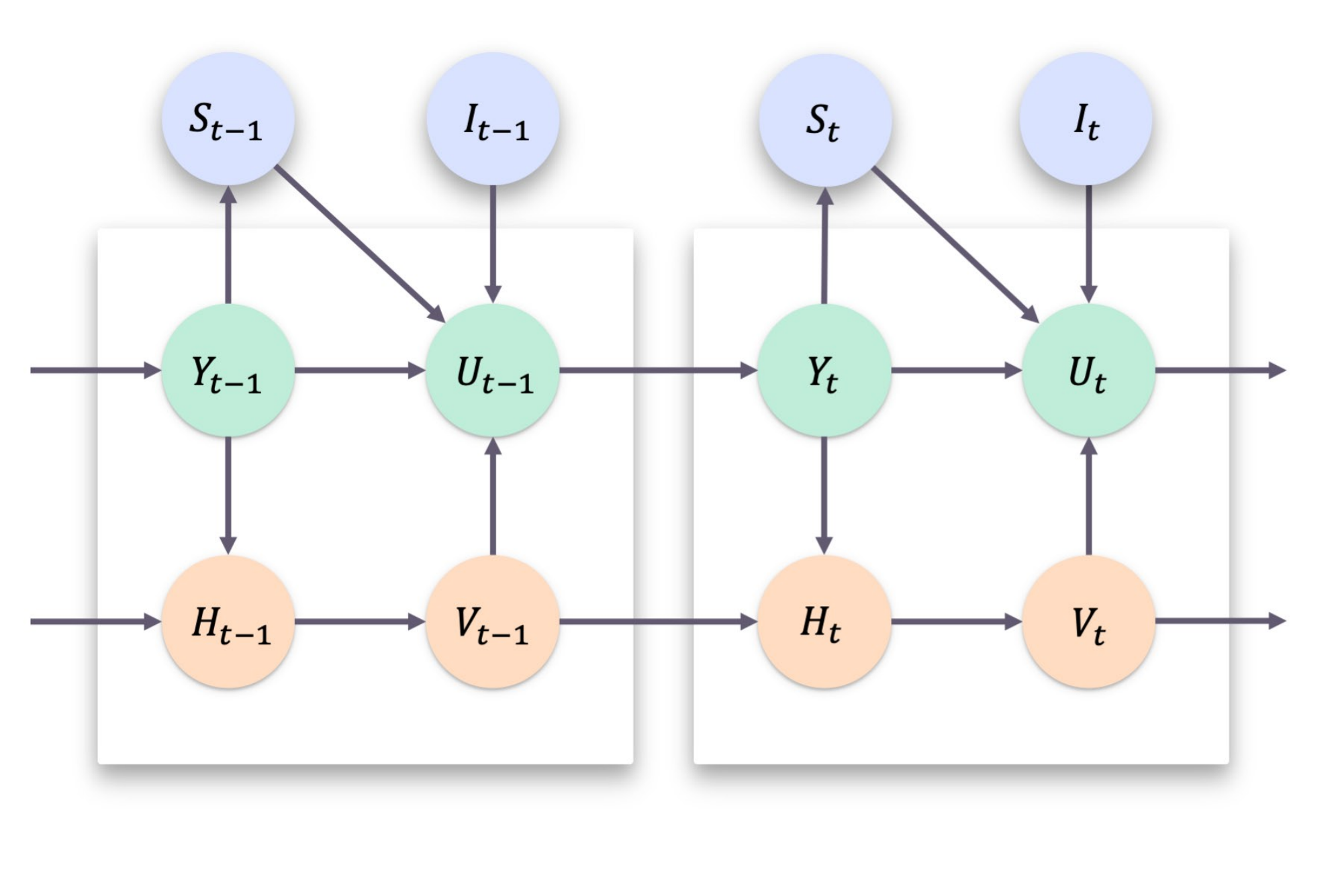}
    \caption{General SIT computational graph.}
    \label{fig:discrete}
    \end{minipage}
\end{figure}

Firstly, We count the input distribution map of the neuron network in Fig.  \ref{fig:distribution}. The input, serves as $I(t)=W\cdot X(t)$, would 90\% distribute between [$-4.92, 3.83$]. Referring to the stable point of the LIF neuron, we expect to set the stable point of the standardized Izhikevich neuron to 0. Before calculation, we fixed $k=1$ for the sake of simplicity. Therefore, in order to put the stable point on $(0, 0)$, we may substitute the above parameters into Eq. \ref{eq:negative_solution} to solve the range of $b$:
\begin{equation}\label{eq: b range}
    b=-1+20I
\end{equation}
Because the range of $I$ has been fully accounted, the range of $b$ can be determined quantitatively in $[-97.4, 75.6]$. As for other parameters $a, c$, we would keep the experience parameters that are stated in the original reference \cite{izhikevich2007dynamical}. Finally, the standardized phase plot is given in Fig. \ref{fig:std}.

For parameter $d$, as the original paper indicates \cite{izhikevich2007dynamical}, bigger $d$ would cause different neuron behaviors because the reset variable would step over the separatrix of the phase plane plot. As the results shown in bionic spike features of SIT neuron of Sect. \ref{sec: type}, the SIT has the capacity of demonstrating different biological property.

\subsection{Surrogate Gradient Descend of Hybrid Neuron Networks}\label{sec:BP}
We take the Surrogate Gradient Descend method as our back propagation training technique. The inputs of spiking neurons are $\boldsymbol I_t^i = \boldsymbol W^{i-1} \boldsymbol X_t^i$ in the $i$-th layer. According to Eq. \ref{std iz}, we use the vectors $\boldsymbol Y_t$ and $\boldsymbol H_t$ to denote the membrane potential and recovery variable after neuronal dynamics at time step $t$, $\boldsymbol S_t^i$ denotes the output spike at time step $t$. From the Fig. \ref{fig:discrete}, we can describe the dynamics of Izhikevich neurons with the following equations:

\begin{equation}
    \begin{aligned}
        & Y_t = U_{t-1} + \frac{k}{\tau}(U_{t-1}- U_{r})(U_{t-1} - U_c) - \frac{1}{\tau}V_{t-1} + \frac{1}{\tau}I_t, \\
        & H_t = V_{t-1} + \frac{a}{\tau}(b(U_t - U_{r})-V_{t-1}), \\
        & S_t = \Theta (Y_t - U_{th}).
    \end{aligned}
    \label{siz_forward}
\end{equation}

Here we set $V_{t} = H_{t}, U_{t} = Y_{t}$. By Fig. \ref{fig:discrete}, we will alternatively update $Y_t$ and $H_t$ gradients at time step $t$.

\begin{equation}
    \begin{aligned}
        & \frac{\partial L}{\partial \boldsymbol Y_t^i} =  \frac{\partial L}{\partial \boldsymbol Y_{t+1}^i} \frac{\partial \boldsymbol Y_{t+1}^i}{\partial \boldsymbol Y_t^i} + \frac{ \partial L_t}{\partial \boldsymbol Y_t^i}, \\
        & \frac{\partial \boldsymbol Y_{t+1}^i}{\partial \boldsymbol Y_t^i} = \frac{\partial \boldsymbol Y_{t+1}^i}{\partial \boldsymbol U_t^i}\frac{\partial \boldsymbol U_t^i}{\partial \boldsymbol Y_t^i} + \frac{\partial \boldsymbol Y_{t+1}^i}{\partial \boldsymbol V_t^i}\frac{\partial \boldsymbol V_t^i}{\partial \boldsymbol Y_t^i},\\
        & \frac{\partial L_t}{\partial \boldsymbol Y_t^i} = \frac{\partial L_t}{\partial \boldsymbol S_t^i} \frac{\partial \boldsymbol S_t^i}{\partial \boldsymbol Y_t^i} = \Theta'(\boldsymbol Y_t^i-\boldsymbol U_{th}^i). \\
    \end{aligned}    
\end{equation}

Then, updating $H_t$ and its gradient chain: 

\begin{equation}
    \begin{aligned}
        & \frac{\partial \boldsymbol Y_{t+1}^i}{\partial \boldsymbol U_t^i} = 1+\frac{k}{\tau}(2\boldsymbol U_t - (\boldsymbol U_{r}+\boldsymbol U_c)), \\
        & \frac{\partial \boldsymbol Y_{t+1}^i}{\partial \boldsymbol V_t^i} = \frac{\partial \boldsymbol Y_{t+1}^i}{\partial \boldsymbol H_t^i}(\frac{\partial \boldsymbol H_t^i}{\partial \boldsymbol U_t^i}\frac{\partial \boldsymbol U_t^i}{\partial \boldsymbol V_t^i}+\frac{\partial \boldsymbol H_t^i}{\partial \boldsymbol V_t^i}),\\
        &\frac{\partial \boldsymbol Y_{t+1}^i}{\partial \boldsymbol H_t^i} = \frac{\partial \boldsymbol U_{t+1}^i}{\partial \boldsymbol V_t^i} = -\frac{1}{\tau},\\
        & \frac{\partial \boldsymbol H_t^i}{\partial \boldsymbol U_t^i} = \frac{\partial \boldsymbol V_t^i}{\partial \boldsymbol U_t^i}= \frac{ab}{\tau}, 
        \frac{\partial \boldsymbol H_t^i}{\partial \boldsymbol V_t^i} = 1-\frac{a}{\tau}, \\
        & \frac{\partial \boldsymbol Y_t^i}{\partial \boldsymbol I_t^i} = \frac{1}{\tau}. \\
    \end{aligned}
\end{equation}

Finally, the learnable parameters can be back propagated by Izhikevich neurons (bias is ignored):

\begin{equation}
    \begin{aligned}
        \frac{\partial L}{\partial \boldsymbol W^{i-1}} =\sum_{t=0}^{T-1} \frac{\partial L}{\partial \boldsymbol Y_{t}^i} \frac{\partial \boldsymbol Y_t^i}{\partial \boldsymbol I_t^i}\boldsymbol X_t^i.
    \end{aligned}
\end{equation}


 \begin{figure}[H]
    \centering
    \includegraphics[width=\linewidth]{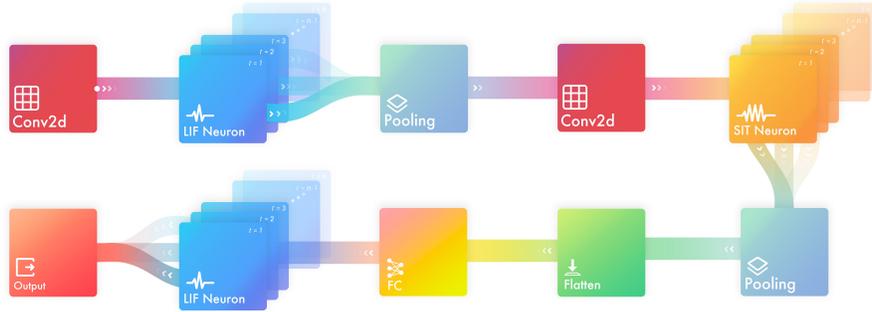}
    \caption{The general formulation of proposed HNN. \textit{FC} denotes the fully connected layer and Pooling represent the spike max-pooling \cite{fang2021incorporating} layer. The faded \textit{LIF neuron} and \textit{SIT neuron} indicate the time series process of spiking neurons. The detailed network architecture of each dataset is shown in Tab. \ref{tab: design}.}
    \label{fig:network arch}
\end{figure}

\section{Experiments}



\subsection{Datasets}
\subsubsection{Datasets Introduction}
The following is the summaries of the datasets and the simulation time step settings utilized in the assessment:

\begin{itemize}
    \item \textbf{MNIST} \ The MNIST dataset is a large collection of handwritten digits, comprised of 28 $\times$ 28 grey-scale image. It has 60,000 examples for training and 10,000 examples for testing, which are labeled from 0 to 9. In MNIST, we set $T = 8$ as the simulation time step.
  
  \item \textbf{Fashion-MNIST} \ Intended to serve as a direct replacement for the original MNIST dataset for benchmarking machine learning algorithms, the Fashion-MNIST dataset contains 28 $\times$  28 grayscale images of 70,000 fashion products from 10 categories, with 7,000 images per category. The training set was consisted by 60,000 images and the test set has 10,000 images. Since the same input channel and pixel scale, we set $T = 8$.

  \item \textbf{CIFAR-10} \ The CIFAR-10 dataset consists of 60,000 32 $\times$  32 images with 3 channels in 10 classes. There are 50,000 training images and 10,000 testing images. For simulation time step, we use the 8 as the time step.

  \item \textbf{N-MNIST} \ The Neuromorphic-MNIST (N-MNIST) dataset is a spiking version dataset converted from MNIST dataset by mounting the ATIS sensor on a motorized pan-tilt unit and moving the sensor while recording MNIST examples on an LCD monitor. It is composed of the same 60,000 training and 10,000 testing samples as the original MNIST dataset. As a neuromorphic dataset, we apply the $T = 10$.
  
  \item \textbf{CIFAR10-DVS} \ The CIFAR10-DVS datasets is the event driven version of the CIFAR-10 dataset. It consists of 10,000 examples in 10 classes, with 1000 examples in each class. Since the CIFAR10-DVS does not divide data into training and testing subsets, we arbitrarily choose 9000 samples as training set and use the rest 1000 samples for testing set by using the same way as \cite{fang2021incorporating,wu2019neunorm} does. We utilize the $T=20$ in preprocess.

  \item \textbf{DVS128 Gesture} \ The DVS128 Gesture dataset is created by event-driven DVS128 camera, including 11 kinds of hand gestures from 29 subjects in 3 different illumination conditions.

\end{itemize}
  
\subsubsection{Dataset Preprocessing}

\begin{figure}
    \centering
    \includegraphics[width=0.5\linewidth]{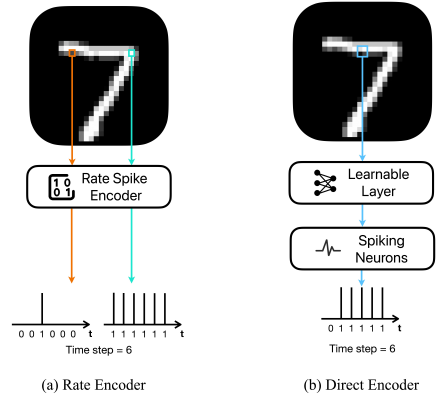}
    \caption{The utilization of both encoders on neuromorphic datasets and static datasets, where (a) Rate Encoder and (b) Direct Encoder, respectively.}
    \label{fig:encoder}
\end{figure}

 In static image datasets, it is necessary to convert static image into spike inputs. there are two major encoding scheme for training large-scale SNNs: rate encoding and neural encoding. The encoding pattern is shown in Fig.\ref{fig:encoder}. The rate encoder encodes the input by generating spike sampled from probability distribution, such as Bernoulli or Poisson distribution. The neural encoder uses a learnable layer to encode static image to spikes. Since the neural encoder is able to achieve higher accuracy \cite{kim2022rate}, We utilize the first $Conv2d - spiking neuron$ as a direct encoder when training HNN in the static image datasets.
 
For details, we leverage the event-to-frame integrating method for neuromorphic datasets preprocessing, which is widely used in SNNs \cite{fang2021incorporating,wu2019neunorm,lisnn2020,kaiser2020decolle}. 

In neuromorphic datasets, the coordinate of event is $E(x_{i}, y_{i}, t_{i}, p_{i})$, event’s coordinate, time and polarity. To reduce the computational consumption, we integrate events into $T$ slices. Note that $T$ is also the simuating time step of HNN. he pixel value is integrated from the events data whose indices are between $j_{l}$ and $j_{r}$:
\begin{equation}
    \begin{split}j_{l} & = \left\lfloor \frac{N}{T}\right \rfloor \cdot j \\
    j_{r} & = \begin{cases} \left \lfloor \frac{N}{T} \right \rfloor \cdot (j + 1), & \text{if}~~ j <  T - 1 \cr N, &  \text{if} ~~j = T - 1 \end{cases} \\
F(j, p, x, y) &= \sum_{i = j_{l}}^{j_{r} - 1} \mathcal{I}_{p, x, y}(p_{i}, x_{i}, y_{i})\end{split}
\end{equation}
$\lfloor \cdot \rfloor$ is the floor operation, and $\mathcal{I}_{p, x, y}(p_{i}, x_{i}, y_{i})$ is an indicator function and it equals 1 only when $(p, x, y) = (p_{i}, x_{i}, y_{i})$.
 
 \begin{table}[H]
\caption{The network architecture setting for each dataset.  \textbf{c}$x$\textbf{k}$y$\textbf{s}$z$/\textbf{MP}$kysz$ is the Conv2D/MaxPooling layer with $output\_channels = x$, $kernel\_size = y$ and $stride = z$. \textbf{DP} is the spiking dropout layer. \textbf{FC}$m$ denotes the fully connected layer with $output\_ feature = m$. \textbf{AP} is the global average pooling layer. the \textbf{\{\}*$n$} indicates the structure repeating $n$ times. }
\label{tab: design}
\resizebox{\textwidth}{0.15\textheight}{\begin{tabular}{@{}ll@{}}
\toprule
Datasets                        & Network Architecture                                                            \\ \midrule
\multirow{2}{*}{MNIST}          & c128k3s1-BN-LIF-MPk2s2-c128k3s1-BN                                              \\
                                & -SIT-MPk2s2-DP-FC2048-LIF-DP-FC100-LIF-APk10s10                                \\ \midrule
\multirow{2}{*}{Fashion-MNIST}  & c128k3s1-BN-LIF-MPk2s2-c128k3s1-BN                                              \\
                                & -SIT-MPk2s2-DP-FC2048-LIF-DP-FC10-LIF                                           \\ \midrule
\multirow{2}{*}{CIFAR-10}       & c256k3s1-BN-LIF-c256k3s1-BN-SIT-c256k3s1-BN                                     \\
                                & -LIF- MPk2s2-\{c256k3s1-BN-LIF\}*3- MPk2s2-DP-FC2048-LIF- DP-FC100-LIF-APk10s10 \\ \midrule
\multirow{2}{*}{N-MNIST}        & c128k3s1-BN-LIF-MPk2s2-c128k3s1-BN                                              \\
                                & -SIT-MPk2s2-DP-FC2048-LIF-DP-FC100-LIF-APk10s10                                 \\ \midrule
\multirow{2}{*}{CIFAR10-DVS}    & c128k3s1-BN-LIF-MPk2s2-c128k3s1-BN                                              \\
                                & -SIT-MPk2s2-\{c128k3s1-BN-LIF-MPk2s2\}*2-DP-FC512-LIF-DP-FC100-LIF-APk10s10     \\ \midrule
\multirow{2}{*}{DVS128 Gesture} & \{c128k3s1-BN-LIF-MPk2s2\}*2-c128k3s1-BN                                        \\
                                & -SIT-MPk2s2-\{c128k3s1-BN-LIF-MPk2s2\}*2-DP-FC512-LIF-DP-FC110-LIF-APk10s10     \\ \bottomrule
\end{tabular}}
\end{table}
 
\subsection{Network Architecture}
 In order to verify the effectiveness of the SIT neuron and compare with the best performance counterpart method fairly, PLIF \cite{fang2021incorporating}, we follow the network architecture of PLIF settings. We propose a general architecture for HNN, which is illustrated in Fig. \ref{fig:network arch}. The spike convolutional layers ($Conv2d - Spiking Neurons$) of HNN are hybridized by LIF neurons and SIT neurons. The detailed network architecture of each dataset is shown in Tab. \ref{tab: design}.

  We use the spiking dropout layer proposed by \cite{lee2020enabling}. Due to the evidently overfitting phenomenon, we fixed the dropout ratio to 0.8 in first dropout layer of CIFAR10-DVS dataset. 
  The dropout ratio for the other dropout layer is 0.5. An additional average pooling layer with $kernel\_size = 10$ and $stride = 10$ dubbed as voting layer, which is implemented after output neurons to alleviate the disturbance of output decoding, except the Fashion-MNIST dataset. 
  The voting layer \cite{wu2019neunorm,fang2021incorporating} is widely used in SNNs to enhance the classify precision with long simulating time step. Since HNN's simulating time step is relatively low, especially the Fashion-MNIST, whose simulating time step is 8. We eliminate the voting layer and directly use the spiking neurons as the output.

\subsection{Parameters tuning}

In our HNN network, we set the Adam optimizer with a 0.01 learning rate and apply the cosine annealing learning rate schedule with $T_{schedule} =64$. The batch size is 16, and the number of iterations for the training step is 1024. Since the process of neuron firing is non-differentiable, we utilize $\sigma(x) = \frac{1}{\pi}\arctan(\pi x) + \frac{1}{2}$ as the surrogate gradient function for back-propagation. 

\begin{table}[h]
\caption{Built-in hyper-parameter of SIT/SIT bursting neuron.}
\centering
\setlength{\tabcolsep}{3.5pt}
\renewcommand\arraystretch{1}
\scalebox{0.78}{
\begin{tabular}{lccccccc}
\toprule
Hyper-parameter & $a$ & $b$ & $c$ & $d$ & $k$ & $U_c$   & $U_{r}$ \\ \midrule
SIT         & 0.002 & 0.02 & 0 & 0.2 & 1 & 1 & -0.05  \\ \hline
SIT Bursting         & 0.35  & 0.6  & 0 & 0.5 & 1 & 1 & -0.05  \\
\bottomrule
\end{tabular}
\label{tab: hyper-parameter}
} 
\end{table}

We empirically set $U_{reset}=0$ and $U_{threshold}=1$ for all neurons. In some prior works \cite{zenke2021remarkable}, the neuronal reset mechanism can be detached from the computational graph to enhance performance, but this is ineffective in HNN. As a result, we maintain the reset process within our computational graph.The membrane time constant $\tau = 2$ is used as the \cite{fang2021incorporating} setting. SIT can conduct a variety of firing modes by altering the built-in hyper-parameter, which is discussed in bionic spike features of SIT neuron in Sect. \ref{sec: type}. Recalling Sect. \ref{sec:izhikevich model} and Eq. \ref{eq: b range}, the hyper-parameter $b$ is in $[-97.4,75.6]$. $a,c$ are the experiment parameters. We choose a reasonably modest $b$ to maintain consistency with the original work \cite{izhikevich2003simple} and bio-interpretability. We raise the value of $d$ from 0.2 to 0.6 to simulate the bursting spike pattern. The settings of built-in hyper-parameter is illustrated in Tab. \ref{tab: hyper-parameter}. Our approach performs optimally when the parameters mentioned above are used.

\begin{table}[!t]
\centering
\caption{Performance comparison between the proposed method and the state-of-the-art methods on different datasets. Best performances are in bold.}
\resizebox{\textwidth}{!}
{
\setlength{\tabcolsep}{3mm}{
\begin{tabular}{lcccccc}

\toprule
\textbf{Model} & {\begin{tabular}[c]{@{}l@{}}MNIST\end{tabular}} & {\begin{tabular}[c]{@{}l@{}}Fashion-MNIST\end{tabular}} & {\begin{tabular}[c]{@{}l@{}}CIFAR-10\end{tabular}} & {\begin{tabular}[c]{@{}l@{}}N-MNIST\end{tabular}} & {\begin{tabular}[c]{@{}l@{}}CIFAR10-DVS\end{tabular}} & {\begin{tabular}[c]{@{}l@{}}DVS128 Gesture\end{tabular}} \\ \midrule
STCA \cite{gu2019stca}   & 98.60\%  & -    & -   & -   & -   & -  \\ \hline
NeuNorm \cite{wu2019neunorm}  & -    & - & 90.53\%  & 99.53\% & 60.50\%  & - \\ 
\hline
ST-RSBP \cite{zhang2019strsbp}  & 99.62\%  & 90.13\%   & -  & - & -  & -   \\ \hline
DECOLLE \cite{kaiser2020decolle} & -   & -   & -  & 96.00\%   & -    & 95.54\%  \\ \hline
LISNN \cite{lisnn2020}    & 99.50\%    & 92.07\%   & -   & 99.45\%   & -   & -  \\ \hline
IIR \cite{fang2020iir}   & 99.46\%  & -    & -   & 99.39\%   & -   & -  \\ \hline
RMP-SNN \cite{han2020rmp}  & -  & -    &\textbf{93.63\%}   & -   & -   & -  \\ \hline
RNL \cite{2021ann2snn}  & 99.46\%  & -    &92.95\%   & -   & -   & -  \\ \hline
Grad R \cite{2021bp}  & 98.92\%  & -    &92.84\%   & -   & -   & -  \\ \hline
PLIF \cite{fang2021incorporating}   & \textbf{99.72\%} & 94.38\%  & 93.50\%  & 99.61\% & \textbf{74.80\%}   & 97.57\%  \\ 
\hline
\hline
\textbf{SIT-HNN} & 99.70\%  & 94.41\%   &  92.93\%  & 99.64\%   & 74.70\%  & 97.57\%\\ \hline
\textbf{SIT Bursting-HNN} & 99.66\%  & \textbf{94.58\%}   &  -  & \textbf{99.66\%}   & 74.60\%  & \textbf{97.92\%} \\ \bottomrule
\end{tabular}}
}
\label{performance}
\end{table}

\subsection{Benchmarks} 

The benchmark consists of one method based one rpresentative biological plasticity rules (\textit{i.e.}, STCA \cite{gu2019stca}), two representative ANN2SNN based methods (\textit{i.e.}, RMP-SNN \cite{han2020rmp}, RNL \cite{2021ann2snn}), and six spike-based back-propagation methods (\textit{i.e.}, NeuNorm \cite{wu2019neunorm}, ST-RSBP \cite{zhang2019strsbp}, DECOLLE \cite{kaiser2020decolle}, LISNN \cite{lisnn2020}, IIR \cite{fang2020iir}, PLIF \cite{fang2021incorporating}, 
Grad R \cite{2021bp}).

For fair comparison, all the compared methods (\textit{i.e.}, NeuNorm, ST-RSBP, DECOLLE, LISNN, IIR and PLIF) are trained on the same training data using Python 3.7.4 with SpikingJelly \cite{SpikingJelly} package based on PyTorch \cite{Pytorch} framework on a workstation equipped with Linux operation system, two Tesla P4 and two Tesla P10 GPUs. As the memory consumption, we use the Tesla P10 to training and testing CIFAR-10 DVS dataset and DVS128 Gesture dataset, and use the Tesla P4 to training and testing other datasets.
\begin{figure}[!t]
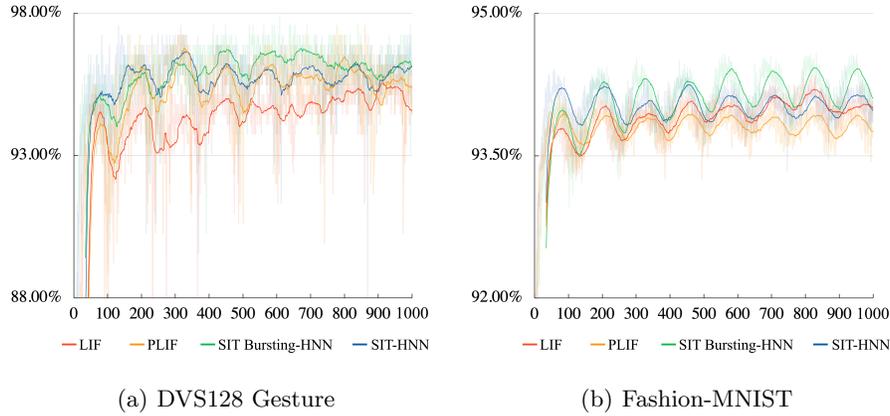

    \centering
    \subfigure[DVS128 Gesture]{
    \label{fig: dvs convergence}
    \begin{minipage}{5.8cm}
    \includegraphics[width=\textwidth]{figures/DVS_convergence.pdf}
    \end{minipage}
    }
    \subfigure[Fashion-MNIST]{
    \begin{minipage}{5.8cm}
    \label{fig: fmnist convergence}
    \includegraphics[width=\textwidth]{figures/fmnist_convergence.pdf}
    \end{minipage}
    }
    \caption{Convergence of the LIF, PLIF, SIT-HNN and SIT Bursting-HNN on different datasets during training. The translucent curve represents the origin data. The solid curves are 35-epoch moving averages. }
    \label{fig: pattern}
\end{figure}

\subsection{Classification Assessment}

The performance of two HNN models is compared with some state-of-the-art (SOTA) models in Tab. \ref{performance}, the HNN is evaluated using two types of SIT neurons, \textit{i.e.}, SIT and SIT Bursting. As shown in Tab. \ref{performance}, the proposed model achieves the SOTA on the N-MNIST, Fashion-MNIST, and DVS128 Gesture datasets. Since only one LIF neuron layer is replaced with an SIT neuron layer, HNN does not reach SOTA performance on each dataset.

\subsection{Discussion}
In this section, we will deeply discuss about the proposed architecture.

\subsubsection{Convergence analysis}
We employ two types of datasets to evaluate the performance of HNN on Fashion-MNIST for static and DVS128 Gesture for neuromorphic. By substituting SIT or SIT Bursting neurons for one LIF neuron layer, our technique achieves faster convergence and higher accuracy.

\begin{figure}[H]
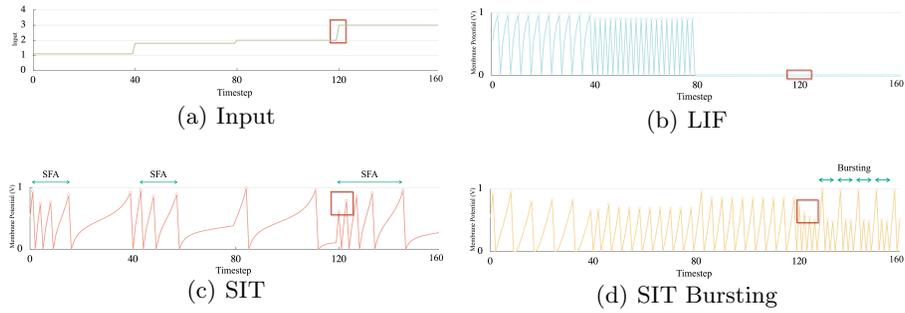

    \centering
    \subfigure[Input]{
    \label{fig: pattern input}
    \begin{minipage}{5.8cm}
    \includegraphics[width=\textwidth]{figures/Input.pdf}
    \end{minipage}
    }
    \subfigure[LIF]{
    \begin{minipage}{5.8cm}
    \label{fig: LIF pattern}
    \includegraphics[width=\textwidth]{figures/LIF_v.pdf}
    \end{minipage}
    }
    \subfigure[SIT]{
    \begin{minipage}{5.8cm}
    \label{fig: SIT pattern}
    \includegraphics[width=\textwidth]{figures/SIT_v.pdf}
    \end{minipage}
    }
    \subfigure[SIT Bursting]{
    \begin{minipage}{5.8cm}
        \label{fig: SIB pattern}
    \includegraphics[width=\textwidth]{figures/Bursting_v.pdf}
    \end{minipage}
    }
    \caption{Spike patterns of SIT, SIT Bursting and LIF when $input = 1.1,1.8,2,3$ and $\tau=2$. Each circle attached to the spike represents a firing. The red rectangles in each figure show the spike pattern conversion receiving the jump of the input.}
    \label{fig: pattern}
\end{figure}

\subsubsection{SIT neuron location analysis}
The location of the SIT neuron is a critical parameter in HNN. We evaluate the performance of the SIT neuron at various positions within the HNN architecture using the DVS128 Gesture dataset and 100 training epochs. As Tab. \ref{tab: design} illustrates, the HNN for classifying DVS128 Gesture consists of 6 convolutional layers and 2 fully connected layers. The SIT location shows where the SIT neuron can be placed in 6 convolutional layers.
To reduce GPU memory consumption, we set the $batch\enspace size = 16$ and use the automatic mixed precision. As shown in Tab. \ref{tab: insert place}, when the SIT neuron in the middle location of network, it achieves higher accuracy.

\begin{table}[H]
\centering
\caption{The table shows the relationship between the accuracy of HNN and the inserted location in 6 convolutional layers.}
\label{tab: insert place}
\setlength{\tabcolsep}{3.5pt}
\renewcommand\arraystretch{1}
\scalebox{0.9}{
\begin{tabular}{lcccccc}
\toprule
SIT location & w/o SIT & 1st & 2nd & 3rd & 4th & 5th \\ \midrule
Accuracy  & 96.18\% & 89.23\% & 94.79\% & \textbf{96.53\%} & 95.83\% & 95.49\% \\ \bottomrule
\end{tabular}
}
\end{table}

\subsubsection{Bionic spike features of SIT neuron}\label{sec: type}

Izhikeivch neuron can model over twenty different types of biological neuron firing pattern \cite{izhikevich2004model}. We choose the tonic spike and the bursting spike as typical neurons for standardizing, and propose the SIT and SIT Bursting neuron. Fig. \ref{fig: pattern} illustrates the firing pattern of SIT, SIT Bursting, and LIF neurons. Compared with LIF Neuron, The SIT neuron exhibits two bionic characteristics:
\begin{enumerate}
    \item \textbf{Spike Frequency Adaption (SFA) \cite{benda2003universal}} \ 
    When receiving a static input, Izhikevich neuron demonstrates reductions in the corresponding firing frequency of their spike response following an initial increase. In biological aspects, the SFA is a common 
    property triggered by inactivation of depolarizing currents and activity-dependent activation of slow hyper-polarizing or shunting currents \cite{benda2003universal}. 
    In our standardized Izhikevich neurons, the SFA property also would come into effect, which is illustrated in Fig. \ref{fig: SIT pattern}.
    \item \textbf{Bursting Firing} \ Observed in neocortex of brain \cite{connors1990intrinsic}, the bursting spiking neuron has higher signal-to-noise ratio than other spiking neurons as the burst threshold is higher than spike threshold \cite{sherman2001tonic}. SIT neurons can fire in bursting mode by adjusting the built-in hyperparameters $a,b,c,d$. When a bursting neuron receives an input, it fires discrete groups repeatedly. Bursting firing mode of SIT Bursting neuron is shown in Fig. \ref{fig: SIB pattern}.
\end{enumerate}
Besides SFA and Bursting, the SIT neuron can fire distinguishable spike pattern (by the different spike frequency) when $input\geq \tau$, which is impossible for LIF neuron. The proof of it can be found in the discussion. As Fig. \ref{fig: pattern} illustrated, the LIF neuron fires the same spike pattern both $input = 2$ and $input =3$.





\subsubsection{Comparison of Spike Encoding Performance.}

\begin{figure}[H]
    \centering
     \subfigure[Input]{
    \begin{minipage}{3cm}
    \includegraphics[width=\textwidth]{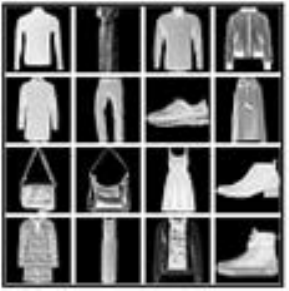}
    \end{minipage}
    }
    \subfigure[SIT]{
    \begin{minipage}{3cm}
    \includegraphics[width=\textwidth]{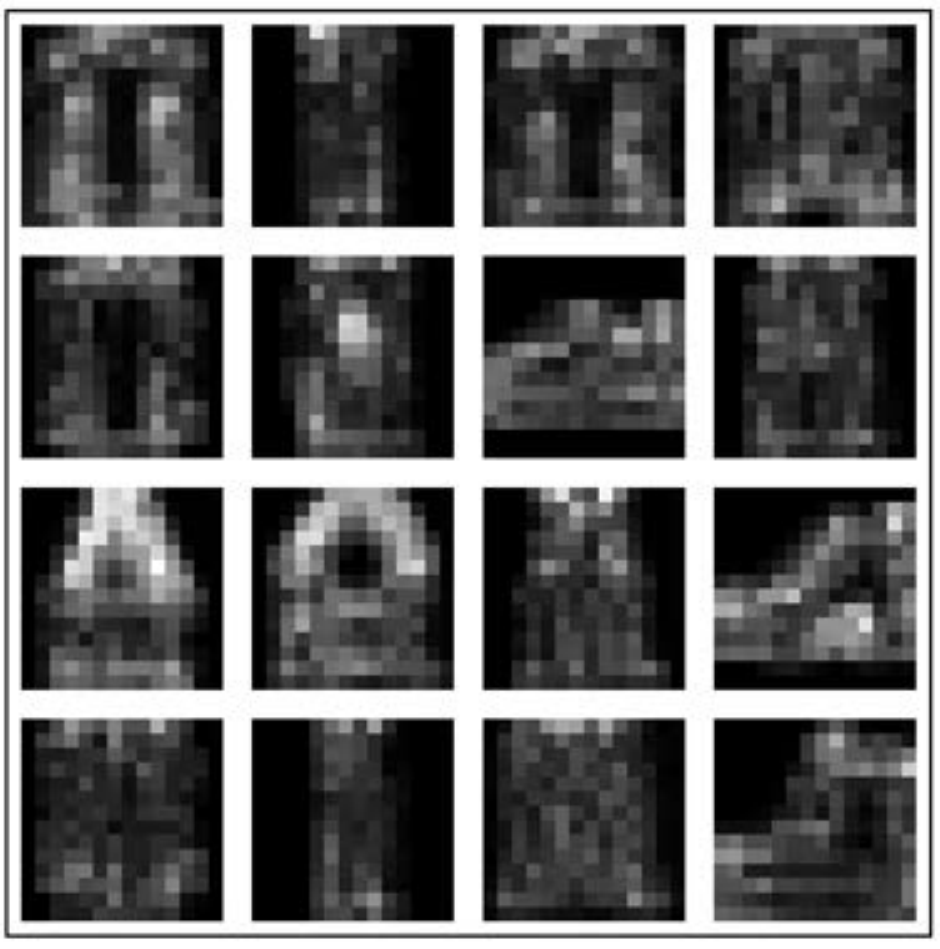}
    \end{minipage}
    }
    \subfigure[LIF]{
    \begin{minipage}{3cm}
    \includegraphics[width=\textwidth]{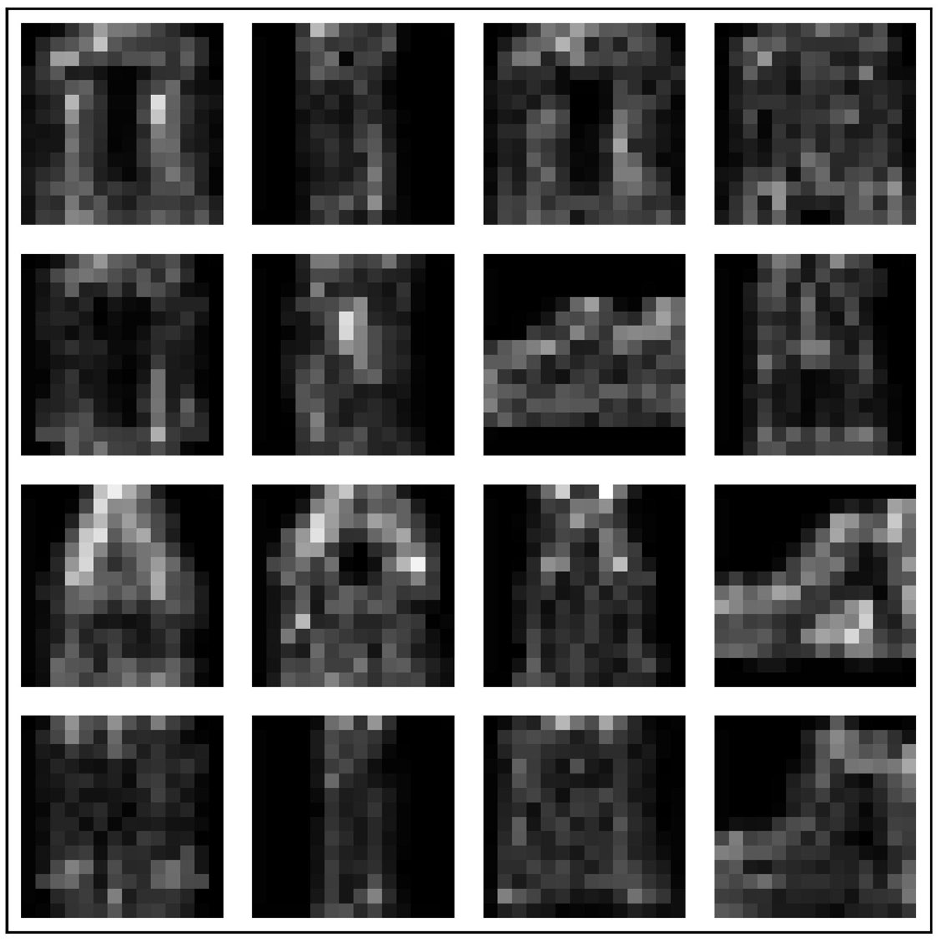}
    \end{minipage}
    }
 
    \caption{Spiking featuremap of SIT and LIF neuron. each column represents a sample and corresponding firing rates.}
    \label{fig: featuremap}
\end{figure}

To evaluate the learning representation performance of SIT neuron, we give 16 samples selected from Fashion-MNIST dataset to get the output spikes $S_{t}^n$ and visualize the accumulation of firing rates $F^n =  \frac{1}{T_s\cdot C}\sum_{c=0}^{C-1}\sum_{t=0}^{T_s-1}S_t^{n}$ as the sample spiking featuremap, where $C$ represent the channels. We compare the SIT neuron with the LIF neuron in the same network position, which is shown in Fig. \ref{fig: featuremap}. As the featuremap illustrate, the spike convolutional layer consist by SIT neuron has a brighter edge than the layer consist by LIF neuron, representing the higher firing frequency in all channel, and the better learning representation.

\subsubsection{The Limitation of LIF Neurons} By revisiting the expression of the LIF neurons, we discover its limitation in the variety of the spiking pattern and proposed the following proposition: 
\begin{proposition}
When $U_{threshold} = 1$ and $U_{reset} = 0$, for any $I_{t} \geq \tau > 1$, the LIF neuron will reach its threshold and reset.
\end{proposition}
\textit{Proof.}  Recall the sub-threshold neuronal dynamics equation of LIF:
\begin{equation}\label{lif equation}
    \tau\frac{dU(t)}{dt} = -(U(t)-U_{reset}) + I(t),
\end{equation}
where $U(t)$ represents the membrane potential of the neuron at time $t$, $I(t)$ represents the input to neuron at time $t$, and $\tau$ is the membrane time constant. In order to align with the discrete setting, the equation of updating the membrane potential should be written as:
\begin{equation}
\label{lif charge}
    U_{t} = Y_{t-1} + \frac{1}{\tau}(-(U_{t - 1} - U_{reset}) + I_{t}),
\end{equation}

\begin{equation}
\begin{aligned}
    \text{when}&\enspace U_{t} \geq U_{threshold},\\ 
    &\enspace Y_{t} = U_{reset},\\ 
    \text{else,}\\
    &\enspace Y_{t} = U_{t},
\end{aligned}
\end{equation}

which describes the membrane potential change in SNN simulation framework.

In the setting of Proposition 1, after the reset, the membrane potential $U_{t}$ will be set to 0. In the next time step, following the Eq. \ref{lif charge} with $U_{r}, U_{t-1}, Y_{t-1}=0$, the membrane potential can be describe as:
    \begin{equation}
    U_{t} = \frac{1}{\tau}I_{t}.
    \end{equation}
As the proposition states, when $I_{t} \geq \tau > 1$,
    \begin{equation}
    U_{t} \geq \tau \geq U_{threshold} = 1.
    \end{equation}
Therefore, the threshold of the LIF neuron is met and it would reset the $Y_{t}$ to 0 and fire a spike, thus resulting in constant spiking.
 
\section{Conclusion}
In this work, we standardize the Izhikevich model in this study using the PPA technique and propose the Standardized Izhikevich Tonic (SIT) neuron. To incorporate the SIT neuron into the SNN, we hybridize it with the LIF neuron in separate convolutional layers and construct the Hybrid Neural Network (HNN). Additionally, we demonstrate that a more complicated spiking neuron can improve the performance of SNNs, which has been shown to be ineffective in prior studies. Due to the early stage of Izhikevich model research, our work contain some limitations: Our entry point into Phase Plane Analysis (PPA) can only solve the parameter $b$ problem quantitatively. For other parameters, their rationale is obscured by computational neuroscience's cross-domain expertise. There fore, additional study is required to overcome the enormous difficulty associated with interpreting the parameters of the Izhikevich model. 

%
%
\bibliographystyle{splncs04}
\bibliography{egbib}

\begin{thebibliography}{10}
\providecommand{\url}[1]{\texttt{#1}}
\providecommand{\urlprefix}{URL }
\providecommand{\doi}[1]{https://doi.org/#1}

\bibitem{akopyan2015truenorth}
Akopyan, F., Sawada, J., Cassidy, A., Alvarez-Icaza, R., Arthur, J., Merolla,
  P., Imam, N., Nakamura, Y., Datta, P., Nam, G., et.al: Truenorth: Design and
  tool flow of a 65 mw 1 million neuron programmable neurosynaptic chip. IEEE
  Transactions on Computer-aided Design of Integrated Circuits and Systems
  \textbf{34}(10),  1537--1557 (2015)

\bibitem{amir2017dvsg}
Amir, A., Taba, B., Berg, D., Melano, T., McKinstry, J., Nolfo, C.D., Nayak,
  T., Andreopoulos, A., Garreau, G., Mendoza, M., et~al.: A low power, fully
  event-based gesture recognition system. In: Proceedings of the IEEE
  conference on computer vision and pattern recognition (CVPR) (2017)

\bibitem{ba2016layer}
Ba, J.L., Kiros, J.R., Hinton, G.E.: Layer normalization. arXiv preprint
  arXiv:1607.06450  (2016)

\bibitem{benda2003universal}
Benda, J., Herz, A.V.: A universal model for spike-frequency adaptation. Neural
  Computation  \textbf{15}(11),  2523--2564 (2003)

\bibitem{bohte2002error}
Bohte, S.M., Kok, J.N., Poutre, H.L.: Error-backpropagation in temporally
  encoded networks of spiking neurons. Neurocomputing  \textbf{48}(1-4),
  17--37 (2002)

\bibitem{adex}
Brette, R., Gerstner, W.: Adaptive exponential integrate-and-fire model as an
  effective description of neuronal activity. Journal of Neurophysiology
  \textbf{94}(5),  3637--3642 (2005)

\bibitem{2021bp}
Chen, Y., Yu, Z., Fang, W., Huang, T., Tian, Y.: Pruning of deep spiking neural
  networks through gradient rewiring. In: International Joint Conference on
  Artificial Intelligence (IJCAI) (2021)

\bibitem{lisnn2020}
Cheng, X., Hao, Y., Xu, J., Xu, B.: Lisnn: Improving spiking neural networks
  with lateral interactions for robust object recognition. In: International
  Joint Conference on Artificial Intelligence (IJCAI) (2020)

\bibitem{connors1990intrinsic}
Connors, B.W., Gutnick, M.J.: Intrinsic firing patterns of diverse neocortical
  neurons. Trends in Neurosciences  \textbf{13}(3),  99--104 (1990)

\bibitem{2021ann2snn}
Ding, J., Yu, Z., Tian, Y., Huang, T.: Optimal ann-snn conversion for fast and
  accurate inference in deep spiking neural networks. In: International Joint
  Conference on Artificial Intelligence (IJCAI) (2021)

\bibitem{fang2020iir}
Fang, H., Shrestha, A., Zhao, Z., Qiu, Q.: Exploiting neuron and synapse filter
  dynamics in spatial temporal learning of deep spiking neural network.
  International Joint Conference on Artificial Intelligence (IJCAI)  (2020)

\bibitem{SpikingJelly}
Fang, W., Chen, Y., Ding, J., Chen, D., Yu, Z., Zhou, H., Tian, Y., et~al.:
  Spikingjelly. \url{https://github.com/fangwei123456/spikingjelly} (2020),
  accessed: 2022-03-02

\bibitem{fang2021incorporating}
Fang, W., Yu, Z., Chen, Y., Masquelier, T., Huang, T., Tian, Y.: Incorporating
  learnable membrane time constant to enhance learning of spiking neural
  networks. In: IEEE International Conference on Computer Vision (ICCV) (2021)

\bibitem{eif}
Fourcaud-Trocm{\'e}, N., Hansel, D., Vreeswijk, C.V., Brunel, N.: How spike
  generation mechanisms determine the neuronal response to fluctuating inputs.
  Journal of Neuroscience  \textbf{23}(37),  11628--11640 (2003)

\bibitem{gerstner2014neuronal}
Gerstner, W., Kistler, W.M., Naud, R., Paninski, L.: Neuronal dynamics: From
  single neurons to networks and models of cognition. Cambridge University
  Press (2014)

\bibitem{gu2019stca}
Gu, P., Xiao, R., G.Pan, Tang, H.: Stca: Spatio-temporal credit assignment with
  delayed feedback in deep spiking neural networks. In: International Joint
  Conference on Artificial Intelligence (IJCAI) (2019)

\bibitem{han2020rmp}
Han, B., Srinivasan, G., Roy, K.: Rmp-snn: Residual membrane potential neuron
  for enabling deeper high-accuracy and low-latency spiking neural network. In:
  Proceedings of the IEEE Conference on Computer Vision and Pattern Recognition
  (CVPR) (2020)

\bibitem{hazan2018bindsnet}
Hazan, H., Saunders, D.J., Khan, H., Patel, D., Sanghavi, D.T., Siegelmann,
  H.T., Kozma, R.: Bindsnet: A machine learning-oriented spiking neural
  networks library in python. Frontiers in Neuroinformatics  \textbf{12}, ~89
  (2018)

\bibitem{stdpizhikevich01}
Heidarpur, M., Ahmadi, A., Ahmadi, M., Azghadi, M.R.: Cordic-snn: On-fpga stdp
  learning with izhikevich neurons. IEEE Transactions on Circuits and Systems
  I: Regular Papers  \textbf{66}(7),  2651--2661 (2019)

\bibitem{hodgkin1952quantitative}
Hodgkin, A.L., Huxley, A.F.: A quantitative description of membrane current and
  its application to conduction and excitation in nerve. The Journal of
  physiology  \textbf{117}(4), ~500 (1952)

\bibitem{izhikevich2003simple}
Izhikevich, E.M.: Simple model of spiking neurons. IEEE Transactions on Neural
  Networks  \textbf{14}(6),  1569--1572 (2003)

\bibitem{izhikevich2004model}
Izhikevich, E.M.: Which model to use for cortical spiking neurons? IEEE
  Transactions on Neural Networks  \textbf{15}(5),  1063--1070 (2004)

\bibitem{izhikevich2007dynamical}
Izhikevich, E.M.: Dynamical systems in neuroscience. MIT press (2007)

\bibitem{lif01}
Jin, Y., Zhang, W., Li, P.: Hybrid macro/micro level backpropagation for
  training deep spiking neural networks. Advances in Neural Information
  Processing Systems (NIPS)  (2018)

\bibitem{kaiser2020decolle}
Kaiser, J., Mostafa, H., Neftci, E.: Synaptic plasticity dynamics for deep
  continuous local learning (decolle). Frontiers in Neuroscience  \textbf{14},
  ~424 (2020)

\bibitem{stdpizhikevich02}
Khoshkhou, M., Montakhab, A.: Spike-timing-dependent plasticity with axonal
  delay tunes networks of izhikevich neurons to the edge of synchronization
  transition with scale-free avalanches. Frontiers in Systems Neuroscience
  \textbf{13}, ~73 (2019)

\bibitem{kim2022rate}
Kim, Y., Park, H., Moitra, A., Bhattacharjee, A., Venkatesha, Y., Panda, P.:
  Rate coding or direct coding: which one is better for accurate, robust, and
  energy-efficient spiking neural networks? arXiv preprint arXiv:2202.03133
  (2022)

\bibitem{koch1998methods}
Koch, C., Segev, I.: Methods in neuronal modeling: from ions to networks. MIT
  press (1998)

\bibitem{lecun1998mnist}
LeCun, Y., Bottou, L., Bengio, Y., Haffner, P.: Gradient-based learning applied
  to document recognition. Proceedings of the IEEE  \textbf{86}(11),
  2278--2324 (1998)

\bibitem{lee2020enabling}
Lee, C., Sarwar, S.S., Panda, P., Srinivasan, G., Roy, K.: Enabling spike-based
  backpropagation for training deep neural network architectures. Frontiers in
  Neuroscience  \textbf{14}, ~119 (2020)

\bibitem{surrogatelif}
Lee, J., Delbruck, T., Pfeiffer, M.: Training deep spiking neural networks
  using backpropagation. Frontiers in Neuroscience  \textbf{10}, ~508 (2016)

\bibitem{li2017cifar10dvs}
Li, H., Liu, H., Ji, X., Li, G., Shi, L.: Cifar10-dvs: an event-stream dataset
  for object classification. Frontiers in Neuroscience  \textbf{11}, ~309
  (2017)

\bibitem{2001lifadaption}
Liu, Y., Wang, X.: Spike-frequency adaptation of a generalized leaky
  integrate-and-fire model neuron. Journal of Computational Neuroscience
  \textbf{10}(1),  25--45 (2001)

\bibitem{maass1997networks}
Maass, W.: Networks of spiking neurons: the third generation of neural network
  models. Neural Networks  \textbf{10}(9),  1659--1671 (1997)

\bibitem{machado2019natcsnn}
Machado, P., Cosma, G., McGinnity, T.M.: Natcsnn: A convolutional spiking
  neural network for recognition of objects extracted from natural images. In:
  International Conference on Artificial Neural Networks (ICANN) (2019)

\bibitem{orchard2015nmnist}
Orchard, G., Jayawant, A., Cohen, G.K., Thakor, N.: Converting static image
  datasets to spiking neuromorphic datasets using saccades. Frontiers in
  Neuroscience  \textbf{9}, ~437 (2015)

\bibitem{Pytorch}
Paszke, A., Gross, S., Massa, F., Lerer, A., Bradbury, J., Chanan, G., Killeen,
  T., Lin, Z., Gimelshein, N., Antiga, L., Desmaison, A., Kopf, A., Yang, E.,
  DeVito, Z., Raison, M., Tejani, A., Chilamkurthy, S., Steiner, B., Fang, L.,
  Bai, J., Chintala, S.: Pytorch: An imperative style, high-performance deep
  learning library. In: Advances in neural information processing Systems
  (NIPS) (2019)

\bibitem{ponulak2010supervised}
Ponulak, F., Kasi{\'n}ski, A.: Supervised learning in spiking neural networks
  with resume: sequence learning, classification, and spike shifting. Neural
  Computation  \textbf{22}(2),  467--510 (2010)

\bibitem{rice2009fpga}
Rice, K.L., Bhuiyan, M.A., Taha, T.M., Vutsinas, C.N., Smith, M.C.: Fpga
  implementation of izhikevich spiking neural networks for character
  recognition. In: International Conference on Reconfigurable Computing and
  FPGAs (ReConFig) (2009)

\bibitem{roy2019towards}
Roy, K., Jaiswal, A., Panda, P.: Towards spike-based machine intelligence with
  neuromorphic computing. Nature  \textbf{575}(7784),  607--617 (2019)

\bibitem{sherman2001tonic}
Sherman, S.M.: Tonic and burst firing: dual modes of thalamocortical relay.
  Trends in Neurosciences  \textbf{24}(2),  122--126 (2001)

\bibitem{lif04}
Shrestha, S.B., Orchard, G.: Slayer: Spike layer error reassignment in time.
  Advances in Neural Information Processing Systems (NIPS)  (2018)

\bibitem{wu2019neunorm}
Wu, Y., Deng, L., Li, G., Zhu, J., Xie, Y., Shi, L.: Direct training for
  spiking neural networks: Faster, larger, better. In: Proceedings of the AAAI
  Conference on Artificial Intelligence (AAAI) (2019)

\bibitem{wu2018group}
Wu, Y., He, K.: Group normalization. In: European conference on computer vision
  (ECCV) (2018)

\bibitem{lif03}
Wu, Y., Deng, L., Li, G., Zhu, J., Shi, L.: Spatio-temporal backpropagation for
  training high-performance spiking neural networks. Frontiers in Neuroscience
  \textbf{12}, ~331 (2018)

\bibitem{xiao2017fashionmnist}
Xiao, H., Rasul, K., Vollgraf, R.: Fashion-mnist: a novel image dataset for
  benchmarking machine learning algorithms. arXiv preprint arXiv:1708.07747
  (2017)

\bibitem{zenke2021remarkable}
Zenke, F., Vogels, T.P.: The remarkable robustness of surrogate gradient
  learning for instilling complex function in spiking neural networks. Neural
  Computation  \textbf{33}(4),  899--925 (2021)

\bibitem{zhang2019strsbp}
Zhang, W., Li, P.: Spike-train level backpropagation for training deep
  recurrent spiking neural networks. Advances in Neural Information Processing
  Systems (NIPS)  (2019)

\end{thebibliography}
\end{document}